\documentclass[lettersize,journal]{IEEEtran}
\usepackage{amsmath,amsfonts}
\usepackage{algorithmic}
\usepackage{algorithm}
\usepackage{array}
\usepackage[caption=false,font=normalsize,labelfont=sf,textfont=sf]{subfig}
\usepackage{textcomp}
\usepackage{stfloats}
\usepackage{url}
\usepackage{verbatim}
\usepackage{graphicx}
\usepackage{cite}
\begin{document}

\title{Dual Swin-Transformer based Mutual Interactive Network for RGB-D Salient Object Detection}

\author{Chao Zeng, Sam Kwong,~\IEEEmembership{Fellow,~IEEE}
\thanks{This paper was produced by Chao Zeng and Sam Kwong. They are with the computer science department of City University of Hong Kong, Kowloon, Hong Kong(e-mail: chao.zeng@my.cityu.edu.hk; cssamk@cityu.edu.hk).}
}

\markboth{Journal of \LaTeX\ Class Files,~Vol.~14, No.~8, August~2021}%
{Shell \MakeLowercase{\textit{et al.}}: A Sample Article Using IEEEtran.cls for IEEE Journals}


\maketitle

\begin{abstract}
Salient Object Detection is the task of predicting the human attended region in a given scene. Fusing depth information has been proven effective in this task. The main challenge of this problem is how to aggregate the complementary information from RGB modality and depth modality. However, conventional deep models heavily rely on CNN feature extractors, and the long-range contextual dependencies are usually ignored. In this work, we propose Dual Swin-Transformer based Mutual Interactive Network. We adopt Swin-Transformer as the feature extractor for both RGB and depth modality to model the long-range dependencies in visual inputs. Before fusing the two branches of features into one, attention-based modules are applied to enhance features from each modality. We design a self-attention-based cross-modality interaction module and a gated modality attention module to leverage the complementary information between the two modalities. For the saliency decoding, we create different stages enhanced with dense connections and keep a decoding memory while the multi-level encoding features are considered simultaneously. Considering the inaccurate depth map issue, we collect the RGB features of early stages into a skip convolution module to give more guidance from RGB modality to the final saliency prediction. In addition, we add edge supervision to regularize the feature learning process. Comprehensive experiments on five standard RGB-D SOD benchmark datasets over four evaluation metrics demonstrate the superiority of the proposed DTMINet method.    
\end{abstract}


\section{Introduction}
The Salient  Object Detection(SOD) task aims to detect the most attended regions for human vision in the testing scene. It has plenty of application scenarios in real life like object segmentation recognition and visual data compression. 
Early salient object detection was mainly based on hand-designed feature extraction methods. In recent years some saliency detection models have been proposed\cite{tang2019salient, wei2020f3net, zhang2021few, li2020learning, zheng2021weakly, zhang2020multi, li2021dense, wang2020deep, mei2021exploring}. Moreover, some works focused on modeling co-saliency relationships are also proposed\cite{fan2021re, zhang2019co, zhang2020adaptive, gao2020co, han2017unified}. However, many of them are based on RGB color images. With the advancement of image acquisition technology, a variety of auxiliary modal image acquisition methods have been proposed, such as infrared images and depth images. Among them, depth images have been a research hotspot in recent years, and some studies have proven that depth information has an excellent complementary effect to traditional color images on saliency prediction\cite{chen2020dpanet, jiang2020cmsalgan, liu2021visual}. As shown in Fig. 1, the flower object appears more distinguishingly in the RGB image than that in the depth image since color clues will give more meaningful guidance for this scene. However, in the second image, the horse object appears better in the depth image since the background objects are better suppressed in the depth image for this scene. From the two example cases, we can see that RGB and depth modality can complement well with each other for the saliency prediction task. In addition, ddge information has been proven helpful for the saliency prediction task in previous work\cite{liu2021visual, jiang2020cmsalgan}. From the figure, we can also find that the edge prediction and saliency prediction are intuitively inter-correlated. So far, although many models for RGB-D saliency detection have been proposed, according to our observations, the task still has the following challenges.

\begin{figure}[htbp]
	\centerline{\includegraphics[width=0.95\columnwidth]{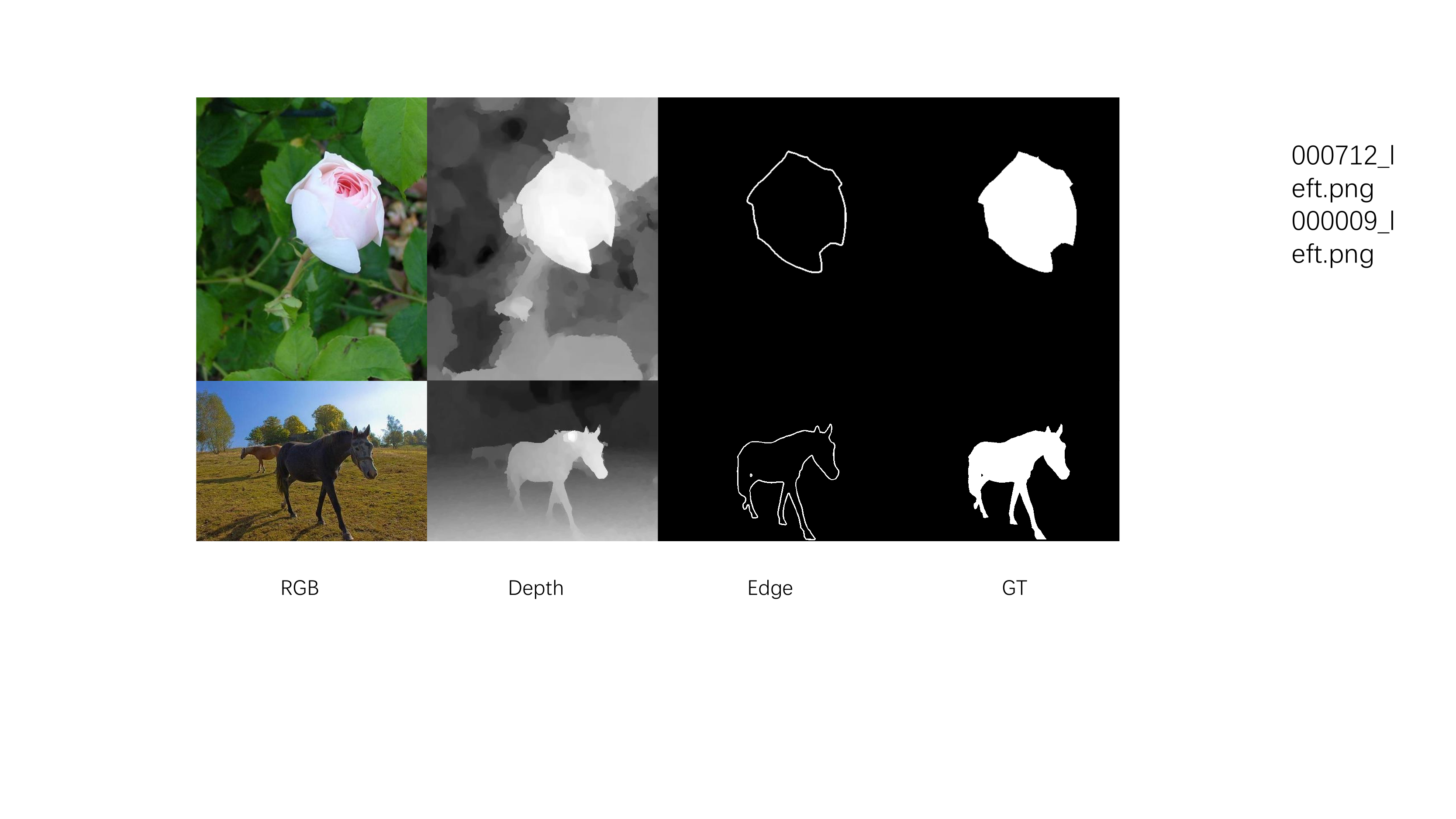}}
	\caption{The illustration of edge regularized RGB-D saliency detection. The RGB modality and depth modality can be complementary with each other. The saliency prediction and edge prediction can co-operate well since they are highly correlated. GT represents ground truth.}
	\label{fig:intro}
\end{figure}

Firstly, compared with traditional RGB SOD tasks, saliency methods incorporating depth information have been less studied. The first model in this field to apply a deep learning method was proposed in 2017, which used low-level CNN features for saliency prediction\cite{qu2017rgbd}. For RGB-D SOD, the deep learning-based method extracts effective features, and there is still a large room for improvement.

Secondly, how to effectively take into account the information of the two modalities of color image and depth, and give play to their respective advantages to improve the performance of the model, is the current research difficulty in the field of RGB-D SOD. Existing deep learning methods can be divided into three categories: pre-aggregation, mid-aggregation and post-aggregation. The pre-aggregation is to merge the input data of the two modalities by splicing and other methods before using the deep network to extract features and then enter the deep model for processing. Post-aggregation refers to first using a deep network to extract the features of the two modalities, fusing the two branches of deep features, and then predicting the saliency map. Other works use middle-level fusion. This fusion method will first use different CNN networks to extract the features of the two modalities and consider the interaction between the two modalities in the feature extraction process.

Thirdly, compared with color images, high-quality depth images are relatively scarce. In the existing datasets, there is a problem that the quality of some depth images is not high, which may lead to the suboptimal solution of the saliency model\cite{chen2020dpanet, zhang2021depth}.

Considering the above challenges faced by the RGB-D SOD task, we propose a dual-stream hybrid encoder-decoder structure based on transformers and CNN. However, unlike existing methods that introduce a separate feature fusion branch like in \cite{liu2021tritransnet}, our method retains the original two-modal dual-stream encoding structure, introduces a transformer-based cross-modality interaction module, and returns the features after the interaction to their succesive stages so that the two modalities have sufficient interaction in the encoding process, and learn the complementary features. Furthermore, to enable the respective encoding branches to learn more effective features, we adopt a feature decoder based on dense connected CNN layers, which effectively supervises both encoding branches. For the quality issue of depth maps, we design the gated modality attention module to filter depth features of higher encoding levels. Besides, we add a skip convolution module to collect the RGB features of early encoding stages for final saliency prediction, which gives more guidance from the RGB modality. We also add an edge prediction head as a regularizer to learn more relevant features for saliency prediction.

In summary, this paper has the following main contributions:
\begin{itemize}
	\item We propose a swin-transformer-based dual-stream mutual interactive network for the RGB SOD task, which we denote as DTMINet. This method uses a dual-stream Swin-Transformer-based structure to encode information from two modalities. Spatial and channel attention are both used to enhance the feature representations of both modalities. The transformer self-attention is used as a bridge to strengthen the interaction of the two branches during the encoding process. A gated modality attention module is applied paired with the interaction module in deeper encoding stages to fuse two branches of features with predicted dynamic weights.
	\item We propose a dense-connected feature decoding structure inspired by DenseNet\cite{huang2019convolutional} to fully supervise the two encoding branches to learn effective features. We also add edge supervision together with saliency supervision, which serves as a regularizer to learn more relevant features.
	\item We compare the proposed DTMINet with 21 state-of-the-art methods on five standard benchmarks for the RGB-D SOD task. In addition, we also conducted a comprehensive ablation experiment to explore settings with the proposed method, such as the input modality data involved, the mode of modal interaction, and the structure of the decoder. Evaluation results based on five standard benchmark datasets show that our proposed method achieves outstanding performances.
\end{itemize}

The rest of this paper is organized as follows. Chapter 2 discusses the related research areas and related methods covered in this paper. Chapter 3 presents the details of the proposed method, followed by Chapter 4, giving the experimental results and related analysis and discussion. Finally, Chapter 5 concludes this paper.

\section{Related Work}
\subsection{RGB-D SOD}
Saliency object detection aims to obtain the most salient regions from the input image. Earlier methods relied on saliency priors and used manual methods to extract features. Although these feature extraction methods are relatively efficient and fast, they may not adapt to complex physical scenes and diverse image content. In pursuit of better generalization, salient object detection based on deep learning has flourished in recent years.
The current mainstream deep learning saliency detection method is the fully convolutional method. To be precise, most of these methods use models represented by VGG\cite{simonyan2014very} and ResNet\cite{he2016deep} as the backbone network to extract features. In order to reinforce the deep features, feature reinforcement methods are usually introduced. One of the most representative is the short-connection method. The core idea of the short-connection method is to transform the size of the depth features at different levels to the same scale and then stitch them together in channel dimension or perform point-by-point operations to merge. Attention mechanisms are commonly used to combine various deep features to obtain superior performance. In addition, the edge information of objects is also shown to be effective auxiliary information to enhance the performance of saliency detection\cite{jiang2020cmsalgan, liu2021visual}.
As another modality, the depth map can give a new perspective to describe the scene from the perspective of the depth of the object physically. For the RGB-D SOD task, the SOD methods will lay the foundation when only the RGB modality is available. The difference lies in how the additional depth map is effectively used to improve the performance of saliency detection compared to the RGB SOD task.
Color images have rich object detail information, such as object color information. But precisely because of this richness of information, it also brings possible ambiguity. For example, the color information of the object makes the deep models prone to mix up the salient objects with background distractors. Models that rely on color images as input can be challenging if the appearance of objects in an image shows a variety of complex color or texture combinations. However, depth information is more robust in this regard, and different regions of the same object usually have depth consistency rather than reflect their apparent differences. Therefore, introducing depth information will undoubtedly bring additional gains to saliency detection. Here, the focus of introducing depth information is how to design the interactive fusion mechanism of two modal data so that the model can obtain more effective saliency-related information.

Currently the contemporary RGB-D SOD methods are basically modeled with CNN layers\cite{chen2019discriminative, pang2020multi, wei2020f3net, fu2020jl, zhang2021depth, wu2021mobilesal, wen2021dynamic, li2021hierarchical, ji2020accurate, zhang2020asymmetric, chen2021cnn, liu2020learning, li2021hierarchical, pang2020hierarchical, li2020icnet, fan2020bbs, wang2020knowing, chen2021depth, liao2020mmnet, chen2021ef, jin2021cdnet, ji2021calibrated, huang2021employing, zhao2021rgb}. Some more recent work also consider utilizing more complex network structure like GAN\cite{jiang2020cmsalgan, tang2019salient, ji2021calibrated} and VAE\cite{zhang2021uncertainty}.
Since the prospering of the transformers\cite{vaswani2017attention}, there are also some works are built based on transformers\cite{wang2021mtfnet, liu2021visual, pang2021transcmd, liu2021tritransnet}. Based on the above considerations, the existing RGB-D SOD methods can be generalized into three categories. Namely pre-fusion, post-fusion, and mid-fusion.

The DTMINet method proposed in this paper belongs to the medium fusion group. However, unlike most existing methods, our method uses transformer layers that are good at capturing long-term dependencies to both extract visual features and fuse the features of the two modalities. The two branches of features after each interaction are returned to succesive layers. With this dual-stream interactive approach, our method provides sufficient space for the feature encoding branch to interact with both modalities. On the usage of edge information, we use it as a regularizer and optimize together with saliency supervision.

\subsection{Transformers Representation}
The transformer model originated from the self-attention mechanism in the NLP field\cite{vaswani2017attention}.
In NLP, there are usually high correlations and dependencies between different words. The self-attention interaction mechanism between word vectors is proposed to model the probabilistic dependencies of words in neighboring contexts. The core idea is to map each word to three representation vectors Query(Q), Key(K), and Value(V) within a sentence. Then it calculates a similarity score for the currently processed word using its Q vector and the K vector of each word, including itself. The V vector of each word is then weighted with this similarity score, and the resulting sum vector is used as a new representation of the current word. The effectiveness of this method has been widely used for reference in the field of computer vision, including ViT\cite{dosovitskiy2020image}, Swin-Transformer\cite{liu2021swin}. In addition some more specialized transformers are later designed for downstream applications like DETR\cite{zhu2020deformable} for object detection, SegFormer\cite{xie2021segformer} for semantic segmentation, CPTR\cite{liu2021cptr} for image captioning, TransT\cite{chen2021transformer} for visual object tracking,  TRIQ\cite{you2021transformer} for image quality assessment.

However, original pure transformers have some limitations. For example, the number of tokens of a fixed length is required, and the number of parameters will increase significantly as the number of layers deepens, which may need more data to train the model entirely. For another example, before and after the transformation of each network layer of the transformer, its feature dimension usually remains unchanged, which brings challenges to some specific tasks. Swin-Transformer\cite{liu2021swin} is proposed with the token merging mechanism so that the resolution of encoded features can be abstracted into a smaller scale progressively at encoding stages, which shares the motivation of the receptive field from CNN.

On the other hand, CNN has good local feature extraction ability. There are some research methods that integrate CNN and transformer like TransUNet\cite{chen2021transunet}, TransFuse\cite{zhang2021transfuse} and Swin-UNet\cite{cao2021swin} in the area of medical segmentation.

Inspired by the complementary advantages of CNN and transformers, our approach combines the advantages of both transformers and convolutional layers to learn better saliency-related features. Specifically, our method utilizes the Swin-Transformer\cite{liu2021swin} as backbone layers to extract visual features. The transformer self-attention is also used for cross-modality interaction while CNN layers are applied to compose feature enhancement modules and saliency decoder.

\section{Methodology}

\begin{figure*}[htbp]
	\centerline{\includegraphics[width=0.92\textwidth]{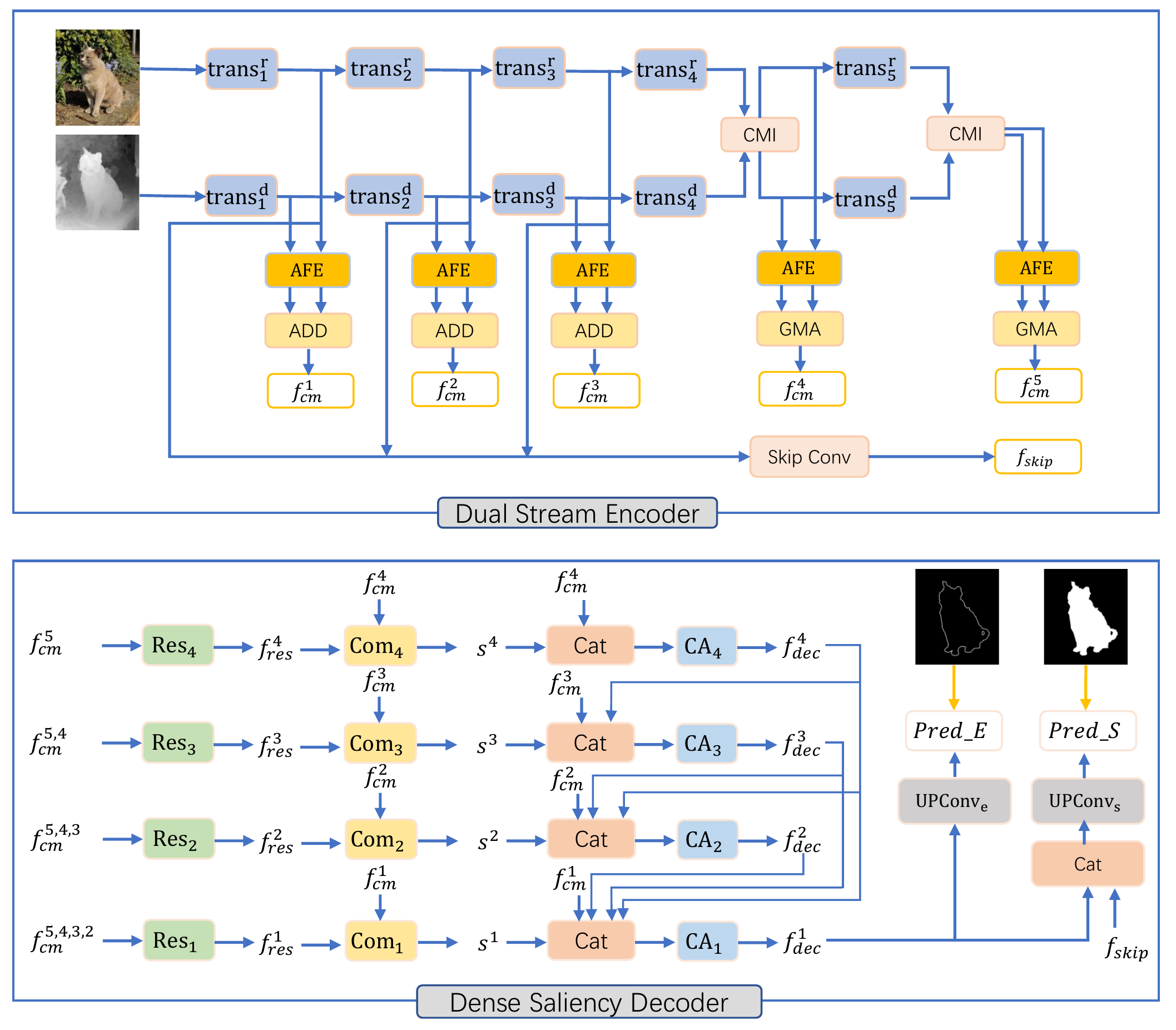}}
	\caption{The Encoder and Decoder Framework of the proposed DTMINet. The ${\rm{trans}}_i^r$ stands for the ${\rm{i}}_{th}$ Swin-Transformer encoding stage for the RGB modality. The AFE stands for Attentive Feature Enhancement. The ADD means addition and production based fusion. The CMI is the Cross Modality Interaction module. The Gated Modality Attention is denoted as GMA. The ${\rm{\{ f}}_{cm}^i\}$ are the multi-level cross-modality features and ${\rm{\{ f}}_{skip}\}$ denotes for the skip features, both of which are the output of the dual Swin-Transformer encoder. There are in total four decoding stages, which are shown in different levels in the figure. The `Res' and `Com' stand for `Residual' and `Common' respectively, which mainly consist of convolutional layers and are illustrated in the DSD section. The `CA' denotes for Channel Attention. `Conv' stands for convolutional layers and `UP' means upsampling. The ${\rm{Pred}}_{E}$ and ${\rm{Pred}}_{S}$ stands for the predicted edge maps and saliency maps, which will be compared to ground truths.}
	\label{fig:encoder}
\end{figure*}


In this section, we present the proposed Dual Swin-Transformer Mutual Interaction Network(DTMINet). In the first subsection, we demonstrate the overview of our method. In the second subsection, we describe in detail the core functional modules in our model, followed by the implementation details in the last subsection.

\subsection{Network Framework}

As shown in Fig. 2, our model adopts the encoder-decoder architecture. It consists of seven modules: RGB Encoder, Depth Encoder, Attentive Feature Enhancement(AFE) module, Cross Modality Interaction(CMI) module, Gated Modality Attention(GMA) module, Skip Convolution(SC) module, and Dense Saliency Decoder(DSD) module.

\textbf{Feature Encoding with Swin-Transformer.} We adopt the dual-stream structure for the feature extraction process for the dual input modalities of RGB and depth images. We use the Swin-Transformer\cite{liu2021swin} as the feature encoding backbone for both RGB and depth inputs. Specifically, the first encoding stage will embed the initial input images into appropriate forms and is followed by four successive Swin-Transformer layers. As shown in Fig. 2, the RGB and depth images are encoded by dual-stream and interactive branches. We enhance the features from RGB and depth modalities with the Attentive Feature Enhancement(AFE) module first and then fuse two branches into one. The CMI and GMA modules are configured in pairs at the last two stages to enhance joint modality representation. Addition and element-wise production are used as fusion methods at early stages while the AFE module is applied on all stages. Considering that there can be inaccurate depth images in the training set, we collect the RGB features of the early three stages into a skip convolution module. The skip features will be considered by the saliency prediction head, which is described in the Dense Saliency Decoder shown in Fig. 2.

\textbf{Cross Modality Interaction.} The CMI module is proposed to make the features from two branches interact. By exchanging information from the two modalities, the dual structured visual feature backbone can extract more modality complementary features to predict salient regions better. After each time of feature interaction, the CMI module will return the two branches of features into the following layers.

\textbf{Gated Modality Attention.} Since salient object detection is a dense prediction task that needs multi-level features to approach good performance, we designed the Gated Modality Attention module to provide multiple connections between encoding layers and decoding layers. The GMA takes the two modality features as input and calculates a gate signal to balance between the two branches of features. Then the fused features are fed into decoding stages as inference clues.

\textbf{Dense Saliency Decoder.} Different from the encoder, We adopt CNN layers in the decoder module. We design several decoding stages and keep decoding histories of intermediate decoding results to make full use of encoding clues. During decoding stages, all previous decoding histories are considered in a progressive manner. In addition, the encoding features from different encoding stages are also merged as additional clues. The RGB features are emphasized with the skip convolution module. We also adopt channel attention for feature enhancement at each decoding stage. Eventually, by taking features both from the decoder and the encoder into account, the saliency map can be predicted from the results of the last decoding stage and the skip features. In addition, an edge prediction head is also applied simultaneously.

\subsection{Swin-Transformer Feature Encoding}
Swin-Transformer is a hierarchically structured transformer network with a window attention mechanism. It introduces window-level attention instead of global attention between every image patch and reduces computational costs by a large margin. In addition, it provides a kind of patch merging mechanism to let the dimension reduce gradually in the encoding process, which imitates the behavior of convolutional encoding layers. We believe the Swin-Transformer combines the merits of both transformer and CNN encoders and employ it as our feature extraction backbone. The feature encoding with the Swin-Transformer can be expressed with the following equations.
\begin{equation}
{\rm{\{ f}}_r^i\}  = swi{n_r}({I_r}),i = 1,2,3,4,5
\end{equation}

\begin{equation}
{\rm{\{ f}}_d^i\}  = swi{n_d}({I_d}),i = 1,2,3,4,5
\end{equation}

In the above equations, the ${I_r}$ and ${I_d}$ are the input RGB and depth images, respectively. The ${\rm{f}}_r^i$ and ${\rm{f}}_d^i$ are the outputs features from the two encoding branches at different stages. There are features from five encoding stages, progressively.
More specifically, given the input image pairs resized as $384 \times 384$, the Swin-Transformer has mainly five encoding stages, which will map the visual inputs into features of various resolutions. Note that the first encoding stage is to shape the input images input proper forms of embeddings without transformer self-attention layers. After the feature encoding process for the input modalities, the model learns to collect the relevant clues for the proceeding saliency reasoning module.

\begin{figure*}[htbp]
	\centerline{\includegraphics[height=6cm]{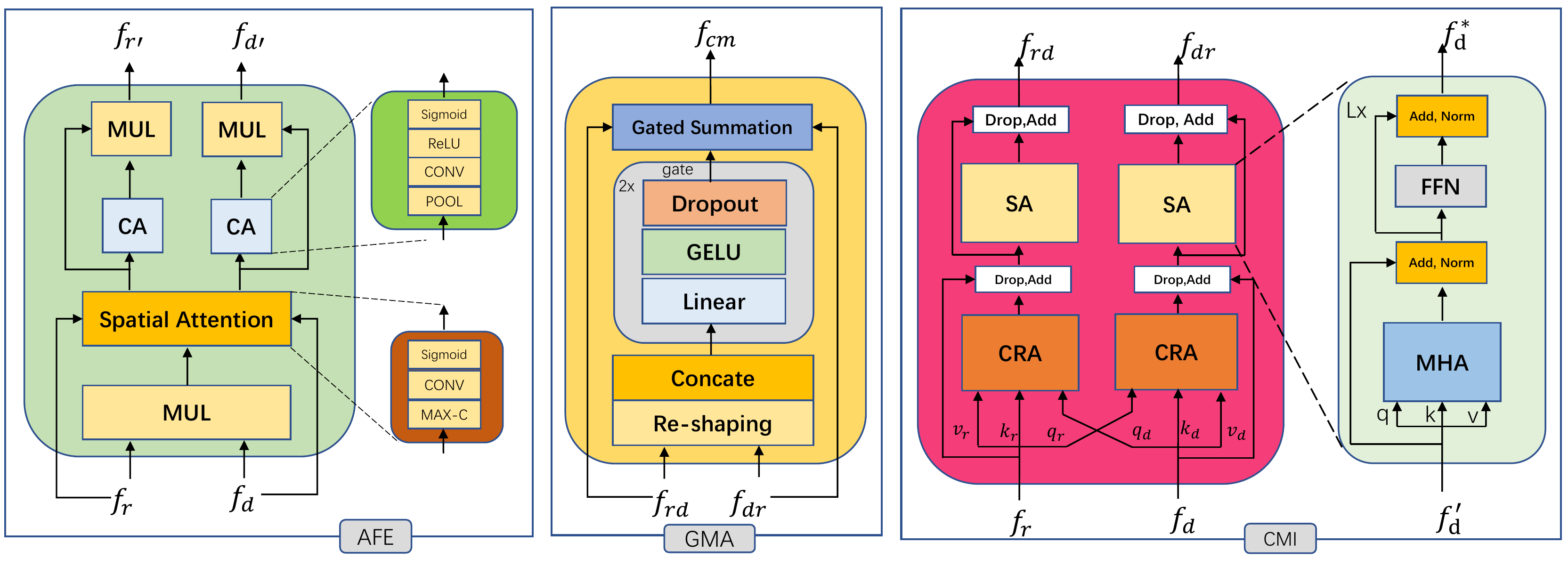}}
	\caption{The detail structures of AFE, GMA and CMI modules. The figure in left shows the inner details of Attentive Feature Enhancement module which combines spatial and channel attention to enhance the RGB and depth features. The `MUL' stands for element-wise multiplication and `CA' denotes for channel attention. The  `MAX-C' in Spatial Attention means taking the max value along channel dimension like max pooling. Similarly the `POOL' operation will erase spatial dimension to focus on channel attention. In the middle, the CMI module is shaped as a stack of cross-modality attention(CRA) and intra-modality self-attention(SA). The `MHA' denotes for multi-head attention of transformers and `FFN' stands for feedforward layers. On contrast to intra-modality self-attention, in the inter-modality cross attention module, the query vectors from two modalities are exchanged for modality interaction. In GMA module, The two branches of modality features are firstly reshaped and concatenated and then passed through the MLP layers to predict a gate signal for weighting the significances of the two modalities. Eventually the gated summation is calculated over the two branches of features to make the cross-modality feature $f_{cm}$.}
	\label{fig:modules}
\end{figure*}

\subsection{Attentive Feature Enhancement}

We utilize an attention module denoted as AFE before fusing them into one combined feature branch of cross-modality feature $\rm{f_{cm}}$. The details of the AFE module are shown in Fig. 3. 

\subsection{Cross Modality Interaction}


In order to associate the two input modalities and leverage the complementary information, we design Cross Modality Interaction(CMI) module based on transformer multi-head self-attention. On the one hand, it serves as a bridge function between the feature encoder and the saliency decoder. On the other hand, it keeps the two-stream structure and returns the interacted features to the original two encoding branches, aiming to enhance the two modality features mutually. The mechanism for the transformer self-attention can be explained with the following equations:

\begin{equation}
{y_0} = [{f_1} + {p_1},{f_2} + {p_2},...,{f_N} + {p_N}],
\end{equation}

\begin{equation}
{q_i} \leftarrow {y_{i - 1}},{k_i} \leftarrow {y_{i - 1}},{v_i} \leftarrow {y_{i - 1}},
\end{equation}

\begin{equation}
y_i^* = LN(MHA({q_i},{k_i},{v_i}) + {y_{i - 1}}),
\end{equation}

\begin{equation}
{o_i} = Soft\max (\frac{{{q_i} \cdot k_i^T}}{{\sqrt {D/{N_h}} }}) \cdot {v_i}
\end{equation}

\begin{equation}
{y_i} = LN(MLP(y_i^*) + y_i^*),i = 1,..,L
\end{equation}
In the above equations, the input features $f$ are firstly formed as N tokens followed by an addition operation with position embeddings. Then the initialized features are mapped into three variables, i.e., the query variable $q_i$, the key variable $k_i$, and the value variable $v_i$. The $\rm{MHA}$ denotes for multi-head self-attention operation, which can be expressed by the equation $(6)$. The D denotes the hidden dimension of the transformer encoder layer, and the $N_h$ stands for the number of heads used by the self-attention module. Followed by further MLP layer and Layer Normalization, the final output for self-attention with the initial feature $y_0$ is $y_i$ at the $i_{th}$ encoding layer. The transformer self-attention can also be explained with the subfigure of CMI in Fig. \ref{fig:modules}. In the Cross Modality Interaction process, the two modalities are firstly processed with cross-modality attention exchanging the query variable in the attention module, which can be expressed with both the following equation and Fig. \ref{fig:modules}. Then the two branches of features are enhanced by self-attention before Add and Drop layer.

\begin{equation}
{y_{rd}} = Soft\max (\frac{{{q_d} \cdot k_r^T}}{{\sqrt {D/{N_h}} }}) \cdot {v_r}
\end{equation}

\begin{equation}
{y_{dr}} = Soft\max (\frac{{{q_r} \cdot k_d^T}}{{\sqrt {D/{N_h}} }}) \cdot {v_d}
\end{equation}
In the above equations, $y_{rd}$ is the RGB feature enhanced by depth modality, and note that it uses the query variable of depth modality. The two equations illustrate the mechanism of the cross-modality attention in the CMI module.

More specifically, the input features for the Swin-Transformer encoder are firstly shaped as [B, L, D]. The B is the batch size, L is the number of tokens that can be taken from the compact dimension fused from the height and width of the corresponding two-dimension feature map. And the D is the hidden dimension of the transformer layers. For each token, it will first be mapped into three representations as Q, K, and V. The similarities between the Q and K vectors are calculated and utilized as weights to summarize the values in vector V, which will be followed by several other proceeding operations like residual connection and normalization. As we can see, with the cross-modality attention in CMI, each modality learns to model contexts with each other. With the following self-attended encoding branch in CMI, the two modalities learns to enhance the individual branch of features to be more contextualized.

\subsection{Gated Modality Attention}

In the previous section, we extend the Swin-Transformer encoding backbone with the Cross Modality Interaction module to enhance the two branches of features with each other. In order to provide the saliency decoder with additional cross-modality clues from the encoding process, we further design a GMA module to combine the two branches of contextualized visual embeddings.

Specifically, we design a gate signal based on  Multi-layer Perceptron and let the network learn to select weights on two modalities according to the input modality embeddings. The mechanism of the GMA module can be expressed with the following equations: 

\begin{equation}
{f_g} = Cat({f_{rd}},{f_{dr}}),
\end{equation}

\begin{equation}
g = GMA({f_g}),
\end{equation}

\begin{equation}
{f_{cm}} = g \odot {f_{rd}} + (1 - g) \odot {f_{dr}},
\end{equation}
in which the $f_g$ is the input feature of the GMA module, which will output a gate signal $g$. The $g$ signal is then utilized for weighting the significance between the two modalities with elementwise production. 

The Gated Modality Attention process can also be explained intuitively in the middle of Fig. 3. The contextualized embeddings $f_{rd}$ and $f_{dr}$ are firstly processed with the re-shaping operation to be shaped as vectors and concatenated together. The combined vector is then processed with two successive fully connected layers, each consisting of a linear layer, the GELU activation function, and the dropout layer. Note that a Sigmoid function is applied before output to normalize the values into the right section. Eventually the predicted weights are utilized to summarize the features from two branches.

\subsection{Skip Convolution}
We add the Skip Convolution module to collect RGB features of early encoding stages as auxiliary clues for saliency prediction, considering the inaccurate depth map issue. It consists of several convolutional layers to perform saliency decoding and can be denoted as the following equation:
\begin{equation}
{f_{skip}} = SkipConv(f_r^1,f_r^2,f_r^3).
\end{equation}
Since the features from different encoding stages are of different scales, the input RGB features are firstly shaped into the same channel dimension with convolutional layers. Successively, for spatial dimention alignment the RGB feature maps are then formed into the same resolution through bilinear interpolation and concatenated by the channel dimension. Channel attention is also applied to enhance the feature representation.

\subsection{Dense Saliency Decoder}

Inspired by the work of DenseNet, we propose the Dense Saliency Decoder to exploit the abundant features from the dual-stream encoder. 

As shown in Fig. 2, we adopt the encoder-decoder structure for the overall framework. There are different decoding levels for the saliency decoder. We apply the dense connections on features from both the encoder backbone and the different levels of decoding stages. The decoding stage that is closest to the encoding bottleneck only has the encoding features, i.e., the  $f_{cm}^5$ and $f_{cm}^4$ from encoder. The ‘cm’ notes for cross-modality aligning to the notation in encoder framework and the number stands for the encoding stage. 

The general goal for the Dense Saliency Decoder is to utilize dense connections to merge the clues from both encoding and decoding stages so it can leverage the saliency decoding performance. After the progressive decoding process, the final decoding feature $f_{dec}^1$ together with skip feature $f_{skip}$ will be utilized to predict the final saliency map. As a regularizer, another edge prediction head is also placed there learning to predict edge map from $f_{dec}^1$. 

Another two significant modules in the dense saliency decoder are the `Res' module and `Com' module, where `Res' stands for Residual and `Com' stands for Common, aiming to mine the residual information among different encoding levels and the contextual relationship information among the encoding features from different stages. The encoding features from the Swin-Transformer backbone at different stages have different resolutions like that in the CNN backbone. The Res module will first map all the input features into the same resolution aligned with the current decoding stages, which can be expressed with the following equation:
\begin{equation}
f_{res}^i = Cat(Up(f_{cm}^{i + 1}),Up(f_{cm}^{i + 2}),...,Up(f_{cm}^5)),
\end{equation}
in which the `Up' denotes for upsampling operation like bilinear interpolation and `Cat' stands for concatenation. The index i denotes the decoding stages, which ranges from one to four. The $f_{cm}^{i}$ denotes the features from the $i_{th}$ encoding stage. After the related encoding features are fused into $f_{res}$, we fuse it with the features from corresponding encoding stage with elementwise product and addition operations, which can be expressed with following equation:
\begin{equation}
f_{com}^i = f_{cm}^i \odot f_{res}^i + f_{cm}^i.
\end{equation}
The product operation will first calculate the common features together with an addition for compact fusion. Note that in Fig. 2, the output of `Com' module is denoted as $s^i$. Here for convenience on notation we denote it as $f_{com}^i$.
The $f_{com}^i$, the encoding feature $f_{cm}^i$ and together with decoding features from all previous decoding stages are then fused. This process is expressed with the following equation:
\begin{equation}
f_{ca}^i = Cat(f_{com}^i,f_{cm}^i,Up(f_{dec}^{i + 1}),...,Up(f_{dec}^4)).
\end{equation}
All these related features are collected and prepared as input into the Channel Attention module, which serves as the final feature enhancement with respect to feature channels. And it can be summarized with the equation:
\begin{equation}
f_{dec}^i = C{A_i}(f_{ca}^i),
\end{equation}
where CA stands for channel attention and $f_{dec}^i$ denotes for the output feature of the decoding stage $i$. Note that the smaller the index for the decoding stage, the closer it is to the final saliency prediction. In other words, the $f_{dec}^1$ will be utilized for final saliency prediction, which is shown with the following equation:
\begin{equation}
s = Up(Conv_{s}(Cat(f_{dec}^1, f_{skip}))),
\end{equation}

\begin{equation}
e = Up(Conv_{e}(f_{dec}^1)).
\end{equation}

Here, the $f_{dec}^1$ is firstly concatenated with skip features and then passed into the saliency prediction head to get the final saliency map $s$. The edge prediction head can be taken as a feature regularizer with additional edge supervision. The skip features are calculated from the RGB modality at early encoding stages, giving more emphasis on the RGB modality and avoiding distractions from the possible inaccurate depth maps.

\subsection{Training Objectives}
On the supervision objective, we utilize the commonly used Binary Cross-Entropy(BCE) loss between the output of the proposed method and the ground truth mask. It can be expressed as the following equations:
\begin{equation}
{{\rm{L}}_s} = \frac{{ - 1}}{{H \times W}}\sum\limits_{i = 1}^H {\sum\limits_{j = 1}^W {[{S_{ij}} \cdot \log G_{ij}^s + (1 - {S_{ij}}) \cdot \log (1 - G_{ij}^s)]} } ,
\end{equation}

\begin{equation}
{{\rm{L}}_e} = \frac{{ - 1}}{{H \times W}}\sum\limits_{i = 1}^H {\sum\limits_{j = 1}^W {[{E_{ij}} \cdot \log G_{ij}^e + (1 - {E_{ij}}) \cdot \log (1 - G_{ij}^e)]} } ,
\end{equation}

\begin{equation}
L = L_{s} + L_{e},
\end{equation}
in which the $G_{ij}^s$ and $S_{ij}$ represent the pixel from the saliency ground truth map and predicted saliency map. The edge loss takes the same form of saliency loss with the difference that the predicted edges are used to compare with the edge labels. The overall optimization loss function L is the summation of the two terms. Compact with all the modules, our proposed model can be trained with the above objectives end-to-end.

\subsection{Implementation Details}
Firstly, we choose the Swin-Transformer encoder and set five encoding stages. And between the deeper encoding layers, we propose the self-attention-based cross-modality interaction module(CMI). The CMI module is implemented as transformer layers. After the modality interaction process, the mutual contextualized features are fed into the succesive layers. The GMA is implemented as a two-layer MLP, in which the GELU activation and dropout layer are used. A Sigmoid function normalizes the gate signal into the range between zero and one. For the saliency decoder, the submodules are implemented with convolutional layers. 

For the edge supervision, we follow VST\cite{liu2021visual} and use edge maps generated from EGNet\cite{zhao2019egnet} for edge labels. Given the input RGB and depth images, the proposed DTMINet will learn to predict saliency and edge maps simultaneously.

We implement the proposed model with PyTorch library and experiment with an NVIDIA 3060 GPU. Following several previous works\cite{zhang2021cross, liu2021tritransnet, liu2021visual}, our method is trained on a composite training set consisting of DUT, NJUD, and NLPR datasets. In the training and testing phase, the input triplets of RGBs, depths, and GTs are first resized into $384 \times 384$ and then normalized. The initial parameters of the feature encoding backbone are inherited from the pre-trained Swin-Transformer\cite{liu2021swin}. We set the embedding dimension for the Swin-Transformer as 128 and the window size as 12. We set the initial learning rate as 5e-5 and a decay rate of 0.1 for every 100 epochs. We choose the Adam optimizer and train our DTMINet with a batch size of 3 for 200 epochs. The average training time for one epoch is about eleven minutes.

\section{Experimental Results}
\subsection{Datasets and Evaluation Metrics}
\textbf{Dataset.} Following the previous work on RGB-D SOD\cite{zhang2021cross, liu2021tritransnet, liu2021visual}, we use a training split which is composed of 1485 triplets of (image, depth, label) from NJUD\cite{ju2014depth}, 700 from NLPR\cite{peng2014rgbd} and 800 from DUT dataset\cite{piao2019depth}. The rest of triplets are reserved for testing. For the model evaluation, we also provide the performance results on SIP\cite{fan2020rethinking} and STEREO\cite{niu2012leveraging}.

\textbf{Evaluation Metrics.}  For performance evaluation, we use MAE, F-max, S-measure, and precision-recall(PR) curve. The four metrics measure the differences between the predicted results and the ground truth masks in various consideration aspects. We use the metrics for both our proposed method and the SOTA methods. And we adopt the evaluation tools from VST\cite{liu2021visual}.
MAE calculates the Mean Absolute Error between the ground truth mask and the predicted map in pixel level:
\begin{equation}
MAE = \frac{1}{{H \times W}}\sum\limits_{i = 1}^H {\sum\limits_{j = 1}^W {|{S_{ij}} - {G_{ij}}|} } ,
\end{equation}
where $S$ and $G$ represent the predicted saliency map and saliency ground truth, respectively. H and W denotes the size of the maps.

The precision rate measures conventional accuracy. Recall measures the ratio of the recalled true positives over all true positives to be predicted:
\begin{equation}
P = \frac{{TP}}{{TP + FP}},
\end{equation}
\begin{equation}
R = \frac{{TP}}{{TP + FN}},
\end{equation}
where TP denotes true positives, FP denotes false positives, and FN stands for false negatives. P and R are precision and recall, respectively.
F-measure combines the precision and recall with a balance parameter $\beta^2$:

\begin{equation}
F = \frac{{(1 + {\beta ^2}) \cdot P \cdot R}}{{{\beta ^2} \cdot P + R}},
\end{equation}
where $\beta^2$ is set as 0.3 following previous works\cite{liu2021visual, chen2020dpanet}. In experiments, we report the maximum F-measure for the predicted saliency maps. S-measure\cite{fan2017structure} combines the object aware structural similarity and region aware structural similarity with a balance parameter $\alpha$. Following previous work the $\alpha$ is set as 0.5\cite{liu2021visual, chen2020dpanet}. In addition, the PR curve of a candidate method can be drawn by setting a series of thresholds on the predicted saliency maps to get the binary prediction map and comparing it to the ground truth map. For each threshold, we can get a pair of precision and recall values, which correspond to one point in the PR curve.
\begin{equation}
{S_m} = \alpha  \cdot {S_o} + (1 - \alpha ) \cdot {S_r}
\end{equation}

\begin{table*}[]
	\centering
	\caption{Performance comparison on NJUD, NLPR, DUT, SIP and STEREO datasets. The comparison methods can be divided into two sections: (1)purly CNN based; (2)transformer based. Among them, VST is fully transformer based from backbone to saliency decoder. For the evaluation metrics of S-m and F-max the higher, the better. MAE is the mean average error between model predictions and ground truths.  }  %
	\label{tab:overall performance}       
	\resizebox{\textwidth}{!}{
		\begin{tabular}{l|lll|lll|lll|lll|lll}
			\hline
			&                & NJUD           &                &                & NLPR           &                &                & DUT            &                &                & SIP            &                &                & STEREO         &                \\ \cline{2-16} 
			Method   & S-m            & F-max          & MAE            & S-m            & F-max          & MAE            & S-m            & F-max          & MAE            & S-m            & F-max          & MAE            & S-m            & F-max          & MAE            \\ \hline
			DMRA\cite{piao2019depth}     & 0.886          & 0.886          & 0.051          & 0.899          & 0.879          & 0.031          & 0.889          & 0.898          & 0.048          & 0.806          & 0.821          & 0.085          & 0.835          & 0.847          & 0.066          \\
			A2dele\cite{piao2020a2dele}   & 0.871          & 0.873          & 0.051          & 0.898          & 0.882          & 0.029          & 0.887          & 0.892          & 0.043          & 0.829          & 0.834          & 0.070          & 0.879          & 0.879          & 0.045          \\
			FRDT\cite{zhang2020feature}     & 0.898          & 0.899          & 0.048          & 0.914          & 0.900          & 0.029          & 0.910          & 0.919          & 0.039          & 0.867          & 0.871          & 0.061          & 0.902          & 0.899          & 0.042          \\
			S2MA\cite{liu2020learning}     & 0.894          & 0.889          & 0.053          & 0.915          & 0.902          & 0.030          & 0.903          & 0.900          & 0.044          & 0.872          & 0.877          & 0.057          & 0.890          & 0.882          & 0.051          \\
			SSF\cite{zhang2020select}      & 0.899          & 0.896          & 0.043          & 0.914          & 0.896          & 0.026          & 0.916          & 0.924          & 0.034          & 0.874          & 0.880          & 0.053          & 0.887          & 0.882          & 0.046          \\
			DANet\cite{zhao2020single}    & 0.897          & 0.893          & 0.046          & 0.909          & 0.894          & 0.031          & 0.890           & 0.895          & 0.047          & 0.878          & 0.884          & 0.054          & 0.892          & 0.882          & 0.047          \\
			CoNet\cite{ji2020accurate}    & 0.894          & 0.893          & 0.047          & 0.907          & 0.887          & 0.031          & 0.919          & 0.927          & 0.033          & 0.858          & 0.867          & 0.063          & 0.905          & 0.901          & 0.037          \\
			D3Net\cite{fan2020rethinking}    & 0.900          & 0.900          & 0.046          & 0.912          & 0.897          & 0.030          & 0.775          & 0.742          & 0.097          & 0.860          & 0.861          & 0.063          & 0.899          & 0.891          & 0.046          \\
			HDFNet\cite{pang2020hierarchical}   & 0.908          & 0.911          & 0.038          & 0.923          & 0.917          & 0.042          & 0.886          & 0.894          & 0.047          & 0.886          & 0.894          & 0.047          & 0.883          & 0.880          & 0.054          \\
			DCF\cite{ji2021calibrated}      & 0.903          & 0.905          & 0.038          & 0.922          & 0.910          & 0.023          & 0.924          & 0.932          & 0.030          & 0.873          & 0.886          & 0.052          & 0.908          & 0.909          & 0.037          \\
			CDNet\cite{jin2021cdnet}    & 0.872          & 0.868          & 0.054          & 0.889          & 0.864          & 0.034          & 0.880          & 0.880          & 0.048          & 0.798          & 0.797          & 0.086          & 0.882          & 0.881          & 0.048          \\
			CDINet\cite{zhang2021cross}    & 0.916          & 0.918          & 0.037          & 0.932          & 0.920          & 0.023          & 0.925          & 0.932          & 0.031          & 0.872          & 0.880          & 0.055          & 0.905          & 0.902          & 0.041          \\
			ACMF\cite{liu2020attentive}   & 0.904          & 0.890          & 0.044          & 0.914          & 0.876          & 0.028          & 0.881          & 0.874          & 0.059          & 0.876          & 0.871          & 0.054          & 0.900          & 0.866          & 0.046          \\
			cmSalGAN\cite{jiang2020cmsalgan}   & 0.903          & 0.897          & 0.046          & 0.922          & 0.907          & 0.027          & -          & -          & -          & -          & -          & -          & 0.900          & 0.894          & 0.050          \\
			CMWNet\cite{li2020cross}   & 0.903          & 0.902          & 0.046          & 0.917          & 0.903          & 0.029          & -          & -          & -          & 0.867          & 0.874          & 0.062          & 0.905          & 0.901          & 0.043          \\
			HAINet\cite{li2021hierarchical}   & 0.909          & 0.909          & 0.038          & 0.921          & 0.908          & 0.025          & 0.916          & 0.920          & 0.038          & 0.886          & 0.903          & 0.048          & 0.909          & 0.909          & 0.038          \\
			RD3D\cite{chen2021rgb}     & 0.916          & 0.901          & 0.036          & 0.930           & 0.892          & 0.022          & 0.931          & 0.924          & 0.031          & 0.885          & 0.874          & 0.048          & 0.911          & 0.886          & 0.037          \\ \hline
			VST\cite{liu2021visual}      & 0.922          & 0.920          & 0.035          & 0.932          & 0.920          & 0.024          & 0.943 & 0.948 & 0.024         & 0.904 & 0.915 & 0.040 & 0.913          & 0.907          & 0.038          \\
			TriTrans\cite{liu2021tritransnet} & 0.920          & 0.919          & 0.030          & 0.928          & 0.909          & 0.020 & 0.933          & 0.938          & 0.025          & 0.886          & 0.892          & 0.043          & 0.908          & 0.893          & 0.033          \\
			MTFNet\cite{wang2021mtfnet}   & 0.921          & 0.923          & 0.032 & 0.932          & 0.924 & 0.020 & 0.937          & 0.946          & 0.024          & 0.898          & 0.906          & 0.040          & 0.906          & 0.906          & 0.039          \\
			TransCMD\cite{pang2021transcmd} & 0.925 & 0.883          & 0.034          & 0.924          & 0.867          & 0.030          & 0.932          & 0.906          & 0.030          & 0.901          & 0.858          & 0.043          & 0.912          & 0.854          & 0.041          \\
			DTMINet  & \textbf{0.929}       & \textbf{0.933} & \textbf{0.029} & \textbf{0.939} & \textbf{0.929}          & \textbf{0.019}          & \textbf{0.944}    & \textbf{0.950} & \textbf{0.021} & \textbf{0.908}          & \textbf{0.918}          & \textbf{0.037}          & \textbf{0.922} & \textbf{0.920} & \textbf{0.031} \\ \hline
	\end{tabular}}
\end{table*}

\subsection{Comparisons with State-of-the-Arts Methods}
In this section we conduct experiments to compare the  performance of the proposed method with various number of current SOTA methods including DMRA\cite{piao2019depth}, A2dele\cite{piao2020a2dele}, FRDT\cite{zhang2020feature}, S2MA\cite{liu2020learning}, SSF\cite{zhang2020select}, DANet\cite{zhao2020single}, CoNet\cite{ji2020accurate}, D3Net\cite{fan2020rethinking}, HDFNet\cite{pang2020hierarchical}, DCF\cite{ji2021calibrated}, CDNet\cite{jin2021cdnet}, CDINet\cite{zhang2021cross}, ACMF\cite{liu2020attentive}, HAINet\cite{li2021hierarchical}, RD3D\cite{chen2021rgb}, VST\cite{liu2021visual}, TriTrans\cite{liu2021tritransnet}, MTFNet\cite{wang2021mtfnet} and TransCMD\cite{pang2021transcmd}. Some of the results are reported from HAINet\cite{li2021hierarchical} and TriTrans\cite{liu2021tritransnet}.

\textbf{Quantitative Comparison.} Table 1 shows the quantitative comparison results over five benchmarks, including NJUD, NLPR, DUT, SIP, and STEREO. We follow the training splits of previous works\cite{zhang2021cross, liu2021tritransnet, liu2021visual} and utilize in total 2985 image triplets from training splits of NJUD, NLPR, and DUT dataset. From the results of Table 1, our proposed method DTMINet achieves the best scores among all five benchmarks and three metrics. From the table, we can also see that transformer-based models generally outperform the CNN-based ones, which validate the effectiveness of transformers applied for the RGB-D SOD task. The PR curves in Fig. 6 also verify the effectiveness of our proposed DTMINet in RGB-D salient object detection. Note that Adele and $\rm{Adele\_s}$ both refer to the A2dele model\cite{piao2020a2dele}. For the A2dele model, we use the saliency maps generated from the RGB stream released by the authors. For the CDINet\cite{zhang2021cross} model, since the released saliency maps are renamed, we regenerate the saliency maps with the released checkpoint model from the authors for the STEREO dataset.

\textbf{Qualitative Comparison.} We further compare the predicted saliency maps as shown in Fig. 4 from a qualitative perspective. The prediction maps of comparison methods are from the released results provided by the authors. Intuitively our proposed method obtains the best visual results compared to six other comparison methods, which are most visually similar to the ground truth masks. Our method is especially good at capturing the detailed characteristics of the salient objects like the butterfly legs for the third case and the occluded back leg of the dog. We can also see that when encountered with distractions from complex backgrounds, our proposed DTMINet can keep the focus on the salient objects. For multi-object scenes, all models need to improve. We visualize the edge prediction results by our method as well paired with saliency results, which are presented in the row of `Edge'. Interestingly, we find our method has a good consistency between the predicted saliency maps and edge maps. It validates that the saliency prediction and edge prediction can co-operate well with each other.

\begin{figure*}[htbp]
	\centerline{\includegraphics[width=0.98\textwidth]{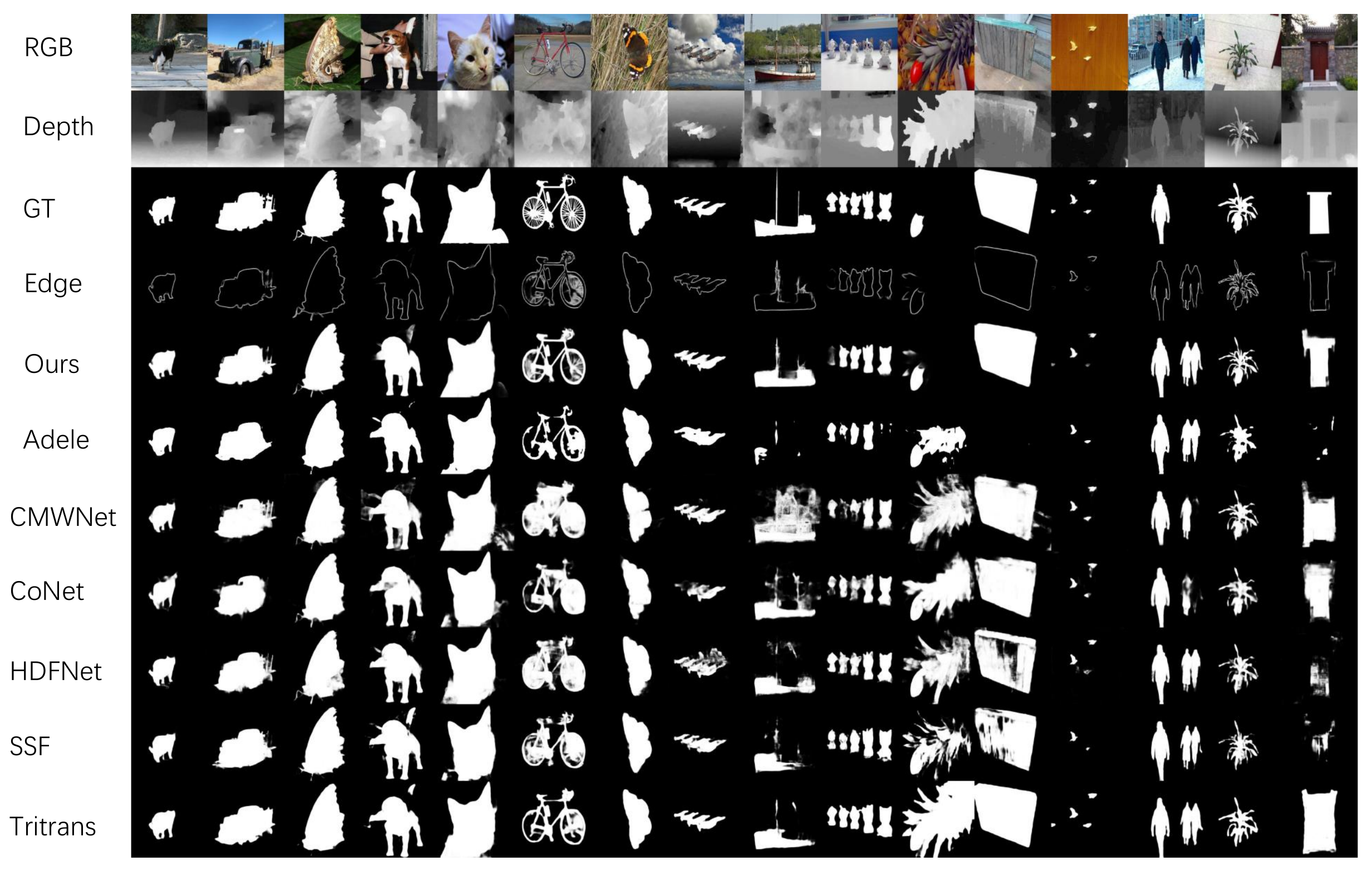}}
	\caption{Visual comparison between state-of-the-art methods and our proposed DTMINet. We also provides predicted edge maps by the proposed method on the testing cases for better comparison, which are presented with the row of `Edge'. }
	\label{fig:vis_one}
\end{figure*}

\begin{figure*}[htbp]
	\centerline{\includegraphics[width=0.98\textwidth]{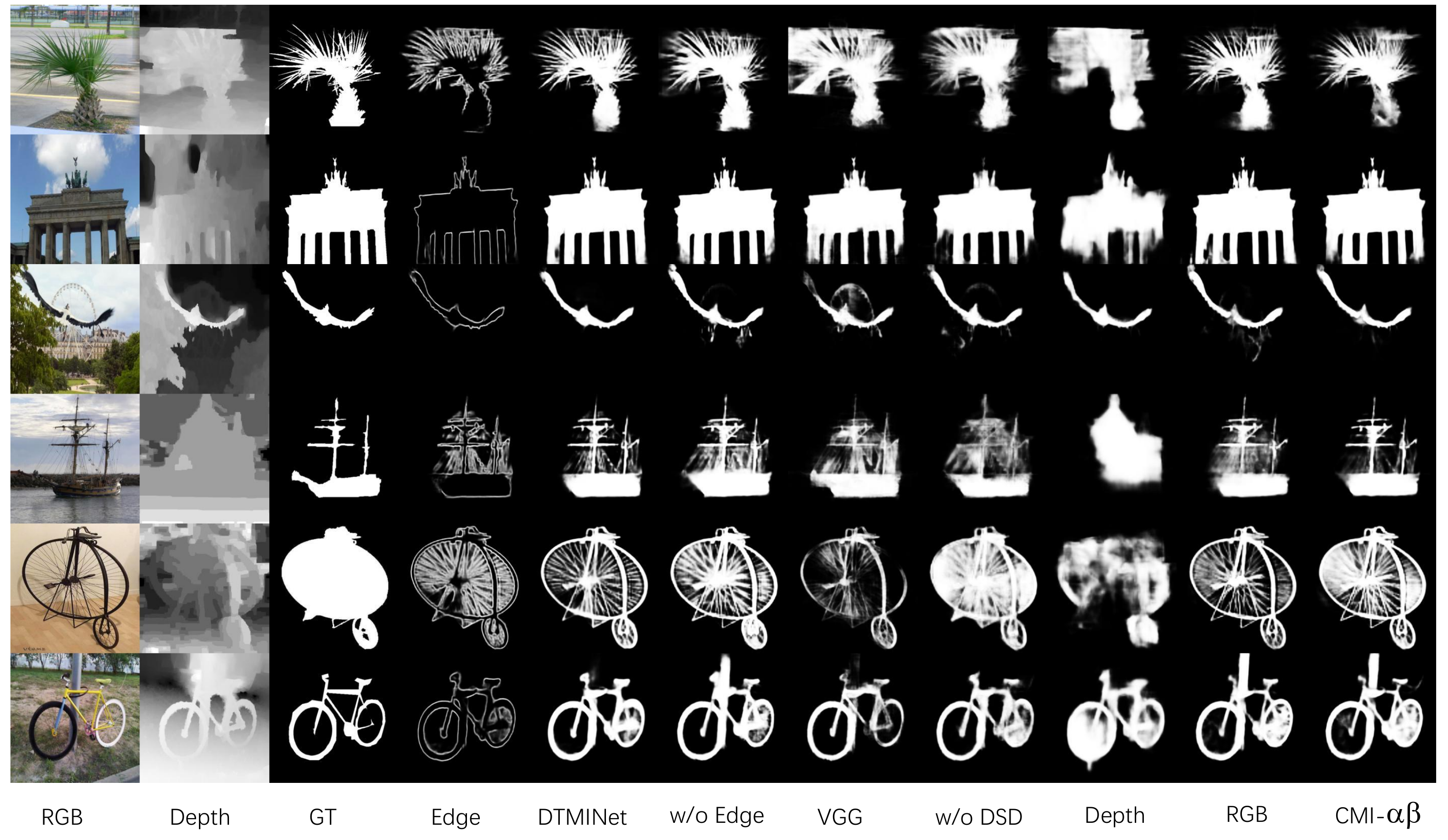}}
	\caption{Visual results of saliency  detection from different variations of our proposed DTMINet. The `Edge' column present the edge prediction results by DTMINet for better comparison.}
	\label{fig:vis_ab}
\end{figure*}


\begin{figure*}[htbp]
	\centerline{\includegraphics[width=0.98\textwidth]{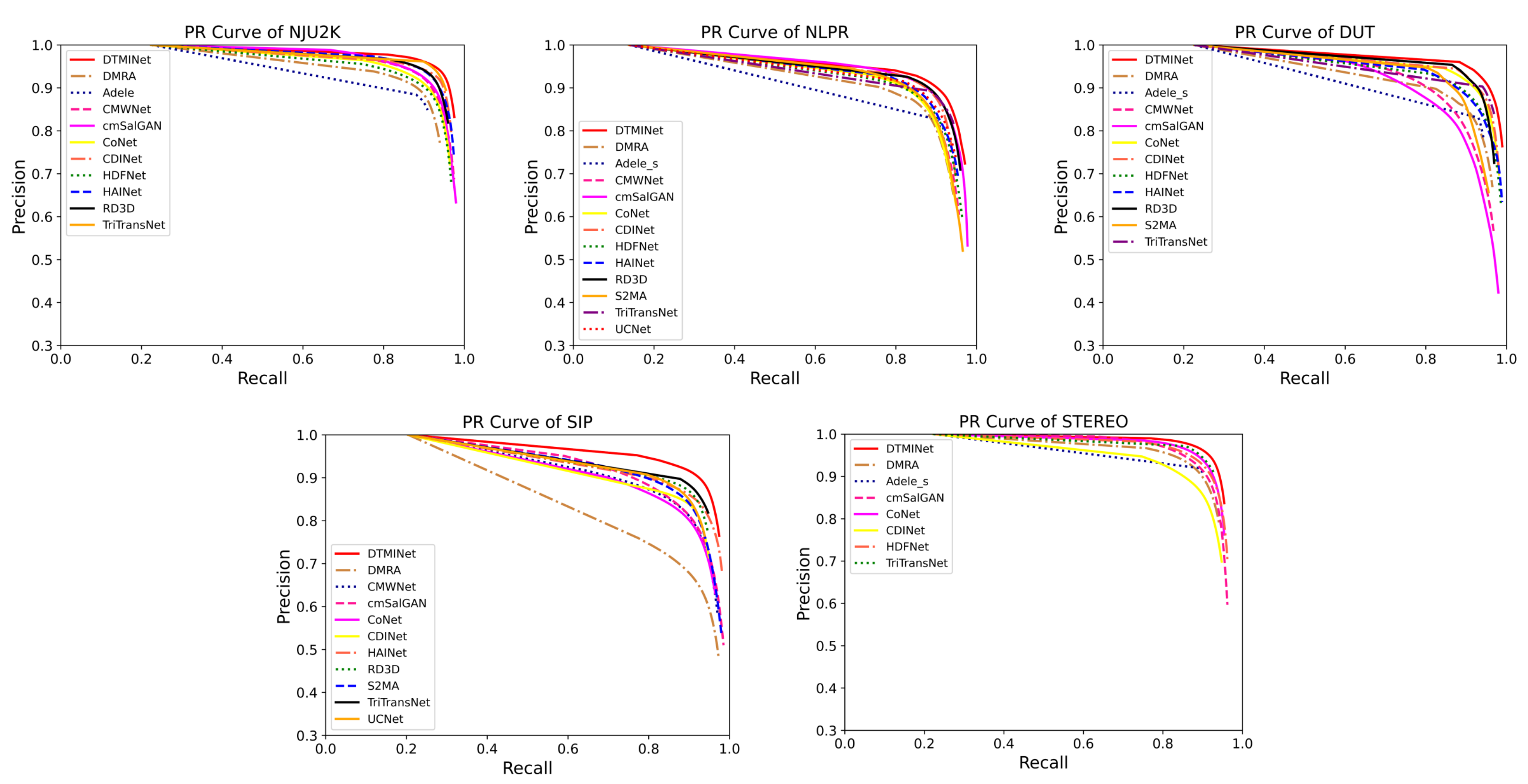}}
	\caption{The PR curves of DTMINet and comparison methods on NJU2K, NLPR, DUT, SIP and STEREO datasets. For comparison methods, we use the saliency maps released by the corresponding authors. It is best viewed in color.}
	\label{fig:pr curve}
\end{figure*}

\begin{table*}[]
	\centering
	\caption{Ablation Studies on NJUD, NLPR and SIP datasets. For the backbone analysis, we replace the Swin-Transformer backbone with VGG16. The w/o F-dec and w/o DSD models are for the decoder analysis. To prove the effectiveness of each modality, we provide single modality models for RGB and depth, respectively. The CMI-$\alpha\beta$ and CMI-V2 models are the variations for the CMI module. In addition, to explore the configurations of applying CMI units to backbone encoding stages, we present the CMI-A, CMI-B and CMI-C models. }  %
	\label{tab:overall performance}       
	\begin{tabular}{l|lll|lll|lll}
		\hline
		&  & NJUD       &       &   & NLPR       &       &    & SIP      &       \\ \cline{2-10} 
		& S-m   & F-max & MAE   & S-m   & F-max & MAE   & S-m   & F-max & MAE   \\ \hline
		DTMINet    & \textbf{0.929} & \textbf{0.933} & \textbf{0.029} & \textbf{0.939} & \textbf{0.929} & \textbf{0.019} & \textbf{0.908} & \textbf{0.918} & \textbf{0.037} \\
		w/o Edge    & 0.925 & 0.927 & 0.031 & 0.937 & 0.926 & 0.021 & 0.892 & 0.902 & 0.046 \\
		VGG16 backbone   & 0.910 & 0.910 & 0.043 & 0.919 & 0.902 & 0.028 & 0.879 & 0.892 & 0.056 \\
		w/o F-dec & 0.923 & 0.922 & 0.033 & 0.933 & 0.921 & 0.022 & 0.883 & 0.889 & 0.052 \\
		w/o DSD    & 0.919 & 0.919 & 0.037 & 0.931 & 0.917 & 0.023 & 0.878 & 0.889 & 0.055 \\
		Depth      & 0.880 & 0.871 & 0.055 & 0.884 & 0.857 & 0.039 & 0.870 & 0.880 & 0.057 \\
		RGB        & 0.922 & 0.925 & 0.033 & 0.936 & 0.925 & 0.020 & 0.883 & 0.894 & 0.052 \\
		CMI-$\alpha\beta$   & 0.924 & 0.926 & 0.033 & 0.934 & 0.924 & 0.021 & 0.876 & 0.884 & 0.055 \\
		CMI-V2 & 0.923 & 0.925 & 0.032 & 0.936 & 0.922 & 0.021 & 0.885 & 0.897 & 0.049 \\ 
		CMI-A    & 0.921 &  0.920  & 0.034 &  0.936 &  0.924 & 0.021 &   0.889  & 0.898 &  0.047 \\
		CMI-B  & 0.924 & 0.924 & 0.033 & 0.936  & 0.925 & 0.021 & 0.881 & 0.888 & 0.053 \\
		CMI-C & 0.921  & 0.921  & 0.034 & 0.932 & 0.919 & 0.023 & 0.891 & 0.900 & 0.048 \\ \hline
	\end{tabular}
\end{table*}

\subsection{Ablation Studies}
In this section, we conduct comprehensive ablation studies on NJUD, NLPR, and SIP datasets to evaluate the compact of the significant modules in our proposed method. To be more specific, we investigate the effectiveness of the key components: a)the effectiveness of transformer encoder representations compared to CNN; b) the importance of combining the two modalities ; c) the effectiveness of the dense connections in decoder; d) the effectiveness of edge information for saliency prediction; e) some variants on the cross-modality interaction module. The results are shown in Table 2. We use both edge supervision and saliency supervision in the full model of DTMINet. For ablation models, only saliency supervision is available.

As backbone analysis, we then implement a CNN-based model as a comparison. The model employs VGG16 as a backbone feature extractor. As shown in Table 2, our method outperforms the model with CNN backbone in all evaluation benchmarks and metrics. The S-measure generally is leveraged by a large margin on every dataset. This validates the effectiveness of the transformer backbone for the saliency prediction task.

Further to demonstrate the effectiveness of the dense salient decoder, we progressively remove the dense connections from decoding histories and encoding contexts, for which the model is denoted with $\rm{w/o F_{dec}}$ and $\rm{w/o DSD}$. As shown in Table 2, when the decoding histories are not considered, the performance drops in all three datasets. However, when the dense saliency decoder is replaced by a plain decoder, which has no dense connection whether from decoding histories or shallow encoding layers, the performance becomes inferior at all datasets, which demonstrates the importance of dense connection in the saliency decoder.

In addition, to further prove the significance of combing both modalities for the RGB-D SOD task, we build the single modality encoder, denoted with RGB and Depth, respectively. From the results, utilizing the RGB modality alone can achieve better performance than with depth modality only. We also observe that only with RGB modality can approach close performance to that of using both modalities on the NLPR dataset. But compared to the NJUD dataset, the performance gap becomes larger. We compared the databases, especially on the RGB images, and found one significant clue. NJUD datasets generally contain images of complex open-world backgrounds with rich and colorful information, which may cause extra challenges for the saliency task. However, the NLPR dataset usually includes images of simpler background like a wall behind the salient object, in which much fewer distractions are present for saliency prediction.

On how to effectively combine the RGB and depth modality, we then provide two variants on the cross-modality interaction module. Firstly, we design a different transformer self-attention-based cross-modality interaction module, denoted by $CMI-V2$. This variant model combines the tokens from both RGB and depth features into a whole representation followed by self-attention modeling. After contextual modeling with self-attention, the universal representations are split into the original two modality encoding branches and continue with the following steps. The results show that this variant model achieves close performances.

Additionally, to further investigate different variations on modality interaction, we refer to \cite{zhang2021depth} and design a variant gating module for the cross-modality fusion, which is denoted as $\alpha\beta$ gate. In the CMI-$\alpha\beta$ variant module, we follow \cite{zhang2021depth} to use the first encoding stage features from both modalities to calculate gate signal $\alpha$. We use the depth features from the last encoding stage to figure out the depth attention map $\beta$. These two kinds of gating signals are used to filter the depth features to send to the RGB encoding branches. Note that no additional supervision is used for learning the gating signals, like depth quality maps. From Table 2, the performance drops slightly for this variant compared to the previous $CMI-V2$ variant on the SIP dataset.

We also did an exploration on the interaction level between the two encoding branches with CMI and GMA modules. The corresponding models are denoted with CMI-A, CMI-B, and CMI-C: a) the version CMI-A is the model with CMI module applied only at stage 5 of the encoder; b) the version CMI-B is the model with CMI modules applied at encoder on stage 5, stage 4 and stage 3; c) the version CMI-C is the model with CMI modules applied at encoder on stage 5, stage 4, stage 3 and stage 2. From the results in Table 2, we can see that when we gradually add the numbers of the stages applied with CMI modules, the performances vary among different datasets. Among them, the proposed DTMINet, which applies the CMI and GMA modules at encoder stage 5 and stage 4, reaches the best performance on all three testing datasets. Generally, the second-best variation model is the CMI-B version, which approaches to closest performances on NJUD and NLPR datasets. 

More over, we conducted a visualization experiment to compare different variational models of our proposed DTMINet in Fig. 5. From the results, we can see that using a single modality will hurt performance. Since high-quality depth images are scarce, only using depth images will cause generally blurry results, as shown in the Depth column. Only using the RGB images, the model tends to suffer from the distractions from complex background objects like in the third and last cases. We can also see that compared to transformer backbone, VGG16 based variation model is less capable of describing object details. The results become relatively coarse when we replace the Dense Saliency Decoder with a simpler structured UNet decoder. For the variation model of CMI-$\alpha\beta$, it can achieve comparable results but is still less optimal for object details like in the first case. Compact with all the submodules, our proposed DTMINet model achieves optimal results while having good consistency with edge predictions.

\section{Conclusion}
In this paper, we propose a hybrid framework for RGB-D salient object detection. Specifically, we employ the Swin-Transformer as the encoding backbone to extract relevant visual features from RGB and depth images framed with a dual structure. Spatial and channel attention is used to enhance the features from two modalities. We further design a transformer-based Cross Modality Interactive module to leverage the cross-modality complementary information and enhance the two branches of features with each other. We propose a Gated Modality Attention module paired with modality interaction to further refine features. We add a skip module to collect RGB features of early encoding stages as additional clues for saliency prediction. We then design a Dense Saliency Decoder to fully explore the abundant clues provided by the dual-stream encoder. The DSD module applies dense connections not only on the encoding features but also on the decoding features. In addition, we also introduce edge supervision as a regularization for learning more relevant features. Both quantitative and qualitative comparisons with contemporary state-of-the-art RGB-D SOD models on various benchmarks show that our proposed DTMINet model achieves outstanding performance.

\section*{Acknowledgments}
This work is supported by Key Project of Science and Technology Innovation 2030 supported by the Ministry of Science and Technology of China (Grant No. 2018AAA0101301), and in part by the Hong Kong GRF-RGC General Research Fund under Grant 11209819  (CityU 9042816) and Grant 11203820 (9042598).

\bibliographystyle{IEEEtran}
\bibliography{refs.bib}
\newpage

\section*{Appendix}

\subsection{Inference Speed Comparison.}
We provide the inference time comparison results in Table 3. We use the test images from the NJUD dataset for testing. In Table 3, we report the average inference time for one testing image. Our DTMINet is implemented with PyTorch. We run the experiment with NVIDIA 3060 GPU. The implementations of other comparison methods are from the originally released versions from the authors. The CDINet model\cite{zhang2021cross} is based on VGG16. The VST model\cite{liu2021visual} is mainly transformer-based. The input image resolution is unified as $384 \times 384$. Originally, the input resolution of CDINet and VST were both $224 \times 224$. We set the resolution of input images as $384 \times 384$ for all testing methods in the speed testing experiment. From the table, we can see that the running speed of our DTMINet is better than the pure transformer-based model VST with or without edge information. Adding the edge module will bring extra computing complexity. The inference speed of DTMINet without edge is close to that of the purely CNN-based model CDINet.
\begin{figure*}[b]
	\centerline{\includegraphics[width=0.98\textwidth]{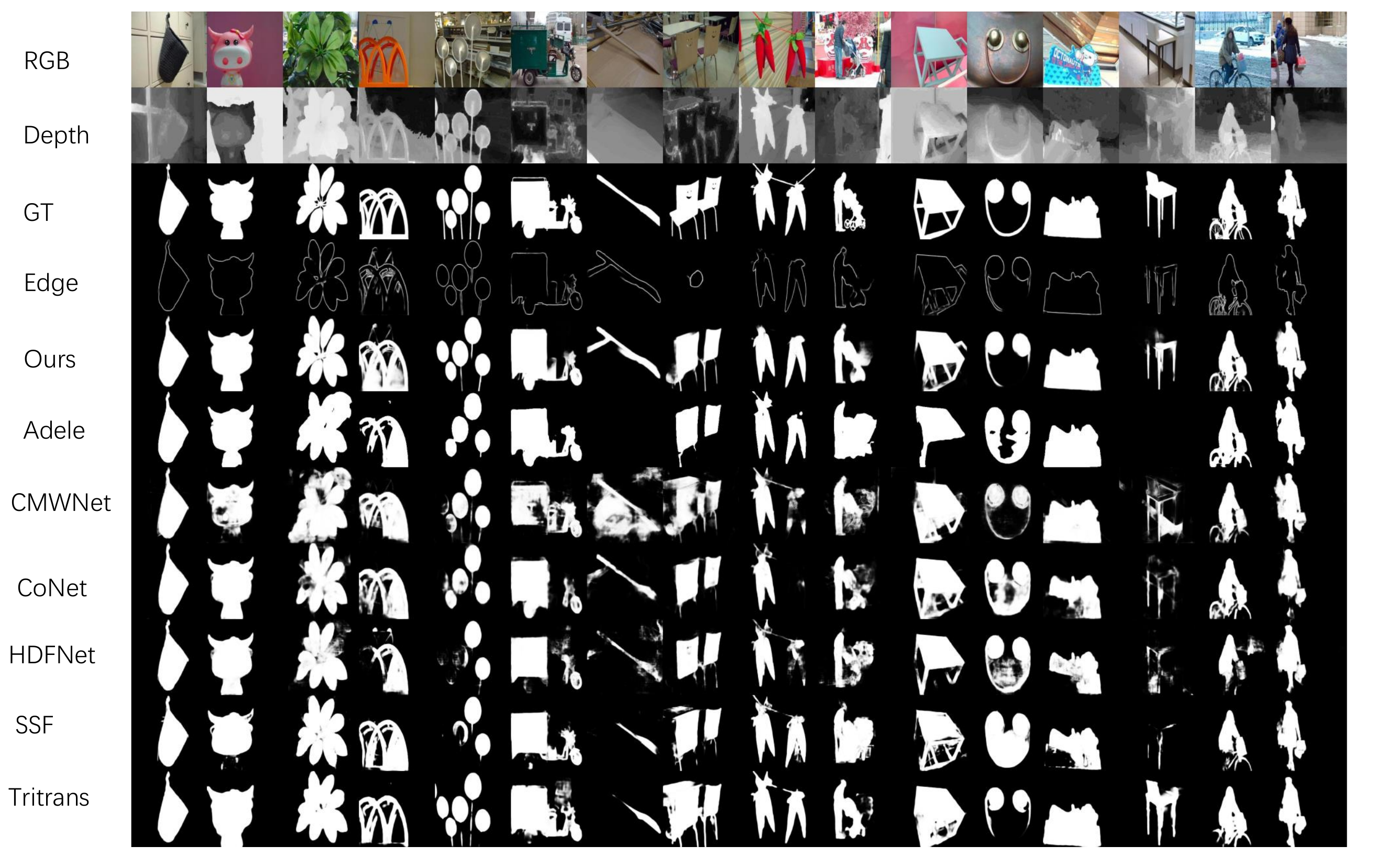}}
	\caption{More visual examples generated by our DTMINet and comparison methods. We also provides predicted edge maps by the proposed method on the testing cases for better comparison, which are presented with the row of `Edge'. }
	\label{fig:vis_app}
\end{figure*}

\begin{table}[b]
	\centering
	\caption{Comparisons of inference time of our proposed DTMINet and comparing SOTA methods with GPU on the NJUD test dataset.}  %
	\label{tab:overall performance}       
	\begin{tabular}{l|llll}
		\hline
		& CDINet\cite{zhang2021cross} & VST\cite{liu2021visual}   & DTMINet w/o edge & DTMINet\\ \hline
		Time (s) & \textbf{0.086}  & 0.149   & 0.090  & 0.116 \\ \hline
	\end{tabular}
\end{table}

\subsection{Additional Visual Examples.}
For better comparison, we then provide with more visual examples in Fig. \ref{fig:vis_app}. Some of the cases are with complex details in the RGB images. Among all the testing methods, our DTMINet can generate relative finer results while showing good consistency with edge prediction.

%
%
%
%

\vfill

\end{document}